\documentclass[letterpaper]{article} % DO NOT CHANGE THIS
\usepackage{aaai2026}  % DO NOT CHANGE THIS
\usepackage{times}  % DO NOT CHANGE THIS
\usepackage{helvet}  % DO NOT CHANGE THIS
\usepackage{courier}  % DO NOT CHANGE THIS
\usepackage[hyphens]{url}  % DO NOT CHANGE THIS
\usepackage{graphicx} % DO NOT CHANGE THIS
\usepackage{natbib}  % DO NOT CHANGE THIS AND DO NOT ADD ANY OPTIONS TO IT
\usepackage{caption} % DO NOT CHANGE THIS AND DO NOT ADD ANY OPTIONS TO IT
\usepackage{algorithm}
\usepackage{algorithmic}
\usepackage{amsmath}
\usepackage{newfloat}
\usepackage{listings}
\usepackage{adjustbox}
\usepackage{multirow}
\usepackage{booktabs}
\DeclareCaptionStyle{ruled}{labelfont=normalfont,labelsep=colon,strut=off} % DO NOT CHANGE THIS
\floatstyle{ruled}
\newfloat{listing}{tb}{lst}{}
\floatname{listing}{Listing}
\title{Uncovering Zero-Shot Generalization Gaps in Time-Series Foundation Models Using Real-World Videos}
\author{
  Lujun Li\textsuperscript{\rm 1},
  Lama Sleem\textsuperscript{\rm 1},
  Yiqun Wang\textsuperscript{\rm 1},
  Yangjie Xu\textsuperscript{\rm 1},
  Niccolò Gentile\textsuperscript{\rm 2},
  Radu State\textsuperscript{\rm 1}
}
\affiliations{
  \textsuperscript{\rm 1}University of Luxembourg \\
  \{lujun.li, lama.sleem, yiqun.wang, yangjie.xu, radu.state\}@uni.lu
  \\
  \textsuperscript{\rm 2}Foyer S.A. \\
  niccolo.gentile@foyer.lu
}

\usepackage{bibentry}
\begin{document}

\maketitle

\begin{abstract}
Recent research on time-series foundation models (TSFMs) has underscored the scarcity of real-world data, often supplemented with synthetic sources in existing datasets, whose generalizability remains however debated. As such, in this work, we propose a novel benchmarking approach: in particular, we aim at building a curated dataset reflecting real world physical temporal dynamics, extracting temporal signals from real-world videos using optical flow. As such, we introduce \textbf{REAL-V-TSFM}, a novel dataset designed to capture rich and diverse time series derived from real-world videos. Experimental results on state-of-the-art TSFMs under zero-shot forecasting show that, despite strong performance on conventional benchmarks, these models exhibit performance degradation on the proposed dataset, suggesting limited generalizability to novel datasets. These findings underscore the need for novel approaches to acquiring time series data and highlight the lack of universality in recent TSFMs, while further validating the effectiveness of our video-based time series data extraction pipeline.
\end{abstract}

\begin{links}
     \small \link{Code}{https://github.com/DobricLilujun/benchmarking_nature_tsfm}
     \small \link{Datasets}{https://huggingface.co/datasets/Volavion/real-v-tsfm}
     \small \link{Extended version}{https://arxiv.org/abs/2509.26347}
\end{links}

\section{Introduction}

\noindent\textbf{TSFM Generalization.} Time series analysis has historically played a central role in practical applications across finance \cite{chen2023chatgpt,yu2023temporal}, healthcare \cite{li2024frozen,liu2023large}, urban computing \cite{wang2023building}, environmental research \cite{dong2023simmtm}, and numerous other fields \cite{nie2024survey}. Foundation Models (FMs) are large pre-trained architectures that learn general patterns from massive data, enabling broad adaptability and strong zero-shot performance across tasks. In natural language processing (NLP), models such as BERT \cite{devlin2019bert} and GPT-3 \cite{brown2020language} have fundamentally reshaped approaches to text comprehension and generation, and the time-series community is undergoing a “BERT moment”, marked by the emergence of foundation Transformer-based models (e.g., Chronos and TSFM variants \cite{ansari2024chronos}). In contrast to NLP, whose generalization performance has been validated by vast numbers of users and researchers \cite{wang2025generalizationvsmemorizationtracing}, the generalization of TSFMs has been far less tested and verified, largely due to limited dataset diversity and a relatively small user base.

\noindent\textbf{Research Question.} In \cite{ansari2024chronos}, the author introduced Chronos, now considered as a cornerstone TSFM. After their evaluation across 42 datasets, both in-domain and zero-shot forecasting, they showed that Chronos surpassed both traditional models and task-specific deep learning approaches. However, the training process mostly rely on synthetic data augmentation\cite{tan2024languagemodelsactuallyuseful,liu2025empoweringtimeseriesanalysis,xie2025caukerclassificationtimeseries}. More specifically, said synthetic training data were generated via two augmentation methods—KernelSynth~\cite{duvenaud2013structure} and TSMixup~\cite{zhang2017mixup}. However, despite employing multiple augmentation strategies, the generalization of these models remains debated. We therefore ask: \textbf{How general can we consider the current TSFMs, and can they really forecast based on real data extracted from daily real events?}

\section{Dataset Construction and Statistics}
\label{dataset_construction}

\subsection{Real World Projection}

To answer our aforementioned research question, we consider specifically time series from video. We chose to use videos since other form of time series like sensor data or stock prices have already been widely investigated in the context of TSFMs. Accordingly, we assess these models with videos as an alternative source of time-series signals. Where appropriate, this strategy can markedly expand available resources for the community, given that videos are among the most abundant time-series data in modern settings. More precisely, camera-recorded video projects 3D scenes onto 2D images, eliminating explicit depth according to the pinhole camera model \cite{sturm2021pinhole}. Videos' high dimensionality and multimodality complicate the extraction of informative univariate signals, yet they still embeds rich temporal patterns that reflect underlying physical dynamics \cite{chari2019visual}. The above technical point once again motivate our choice to leverage existing videos to enrich to assess existing TSFMs' generalizability skills in real-world.

\subsection{Optical Flow Mindset}
\label{sec:optical_flow_mindset}

As a starter, in this paper, we build REAL-V-TSFM, a novel time-series dataset entirely derived using optical flow methods from existing video data. The ultimate goal of optical flow methods is to estimate the motion information of objects within a scene in an image by analyzing changes in pixel intensities over consecutive frames \cite{fleet2005opticalflow}. This method relies on the brightness constancy assumption, which suggests that a moving point in a scene preserves its pixel intensity across adjacent frames \cite{horn1981determining}. When dealing with a large number of diverse long-duration videos, the motion of the main objects within the video frames forms continuous motion patterns. For example, a person on a swing exhibits distinct temporal motion sequences for the hands, waist, and head. Each sequence can be represented along the x- and y-axes, and treated either individually (univariate) or jointly as a multivariate time series. Additionally, camera shake can introduce movements in the background, resulting in a multitude of continuous time series. To our knowledge, this work is the first to propose the use of optical flow through the Lucas-Kanade method \cite{lucas1981iterative} to extract time series signals from pixel trajectories in videos, with particular emphasis on key points. In the following section, we go more in details of the developed workflow to build REAL-V-TSFM.

\subsection{Dataset Production Workflow}

\begin{figure*}[!htbp]
    \centering
    \includegraphics[width=1.0\linewidth]{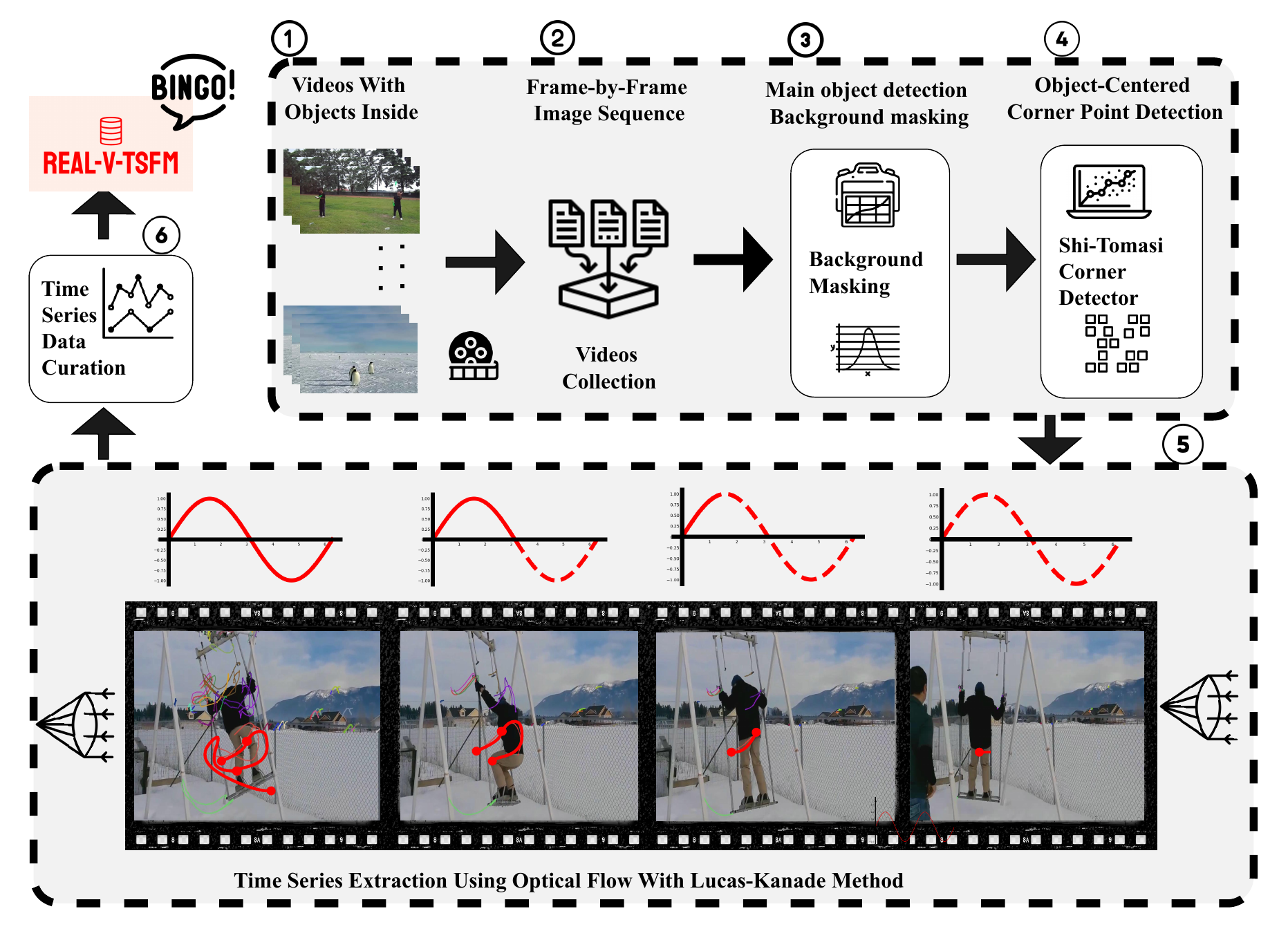}
        \caption{Dataset production workflow consisting of six steps}
    \label{fig:workflow}
\end{figure*}

As a starter, LaSOT \cite{fan2019lasot} serves as the primary video source, offering long sequences with guaranteed main subjects (e.g., humans, animals). As shown in Fig.~\ref{fig:workflow}, videos are first selected \textbf{(Step 1)} and extracted as frame-by-frame images \textbf{(Step 2)}. Foreground detection then uses Mixture of Gaussians 2 (MOG2) \cite{bouwmans2008background,stauffer1999adaptive}, a GMM-based method \cite{han2005update} that models each pixel’s color distribution with multiple Gaussians: pixels not matching background models are classified as foreground \textbf{(Step 3)}. The resulting mask is applied to suppress background, after which corner detection is performed on subjects \textbf{(Step 4)}. Shi–Tomasi's algorithm \cite{shi1994good} is then adopted, identifying corners via a large minimum eigenvalue of the local structure matrix, ensuring strong gradients in both directions. 

Finally, a forward–backward consistency check \cite{kalal2010forward} is performed using pyramidal Lucas–Kanade optical flow to filter unstable trajectories (Step 5). This step removes unreliable trajectories, retaining only correspondences that are consistently tracked across frames. The rationale for this step is that, in our experiments, numerous tracking errors were observed, such as target point loss and misidentifications across multiple frames. Applying the forward–backward consistency check significantly reduces these issues, although a certain amount of noise remains unavoidable.

The forward–backward check is defined as:
\begin{equation}
e_{fb}(\mathbf{p}_0) 
= \left\| \mathbf{p}_0 - \left( f_{\text{backward}} \circ f_{\text{forward}} \right)(\mathbf{p}_0) \right\|_2 ,
\end{equation}
where \(\mathbf{p}_0\) stands for the original pixel (or keypoint) in the first frame
and \( e_{fb}(\mathbf{p}_0) \) denotes the forward-backward error, derived by calculating the Euclidean distance between forward optical flow followed by backward optical flow and comparing the result with the original point. The forward optical flow \( f_\text{forward} \) estimates pixel displacement from the first frame to the second frame, while the backward optical flow \( f_\text{backward} \) estimates the displacement from the second frame back to the first. If \( e_{fb}(\mathbf{p}_0) < \epsilon \), the tracking of point \( \mathbf{p}_0 \) is considered valid, that is to say, the forward and backward tracking is consistent, so the error will be small (close to 0).

In the subsequent post-processing step (\textbf{Step 6}), track durations are standardized by linearly interpolating each track to match the length of the longest track within the same video. Despite background masking, camera motion can introduce background corner tracks; therefore, correlations among all tracks are computed and the five least correlated are retained (subjects typically map to one–two corner points), emphasizing informative, diverse motions while suppressing noise. Finally, each track’s x- and y-coordinates are treated as two independent time-series and stored in REAL-V-TSFM as separate streams but we also propose a multi-variant version. All the technical details are shown in the Appendix.

\subsection{Datasets Statistics}

The dataset contains 6,130 time series corresponding to 609 different objects, reflecting substantial categorical diversity. The series vary markedly in length, with an average of 2,043 time steps (i.e., frames) and a coefficient of variation of 0.516 indicating moderate relative dispersion. The average  sequence lengths of obtained dataset range from 1,000 to 8,000, reflecting a broad variation in temporal resolution across the dataset. Time series values span a broad dynamic range reflecting the positional information in each frame of videos (mean = 402.13, standard deviation = 281.46) highlights pronounced variability across series. Each time series additionally includes the primary object category (e.g., airplane, boat, cat), derived from the corresponding original video dataset. Together, these characteristics underscore the dataset’s rich heterogeneity, making it well-suited for comprehensive time series analysis.

For a comparison with the M4 dataset, the Augmented Dickey–Fuller (ADF) \cite{mushtaq2011augmented} unit root test is applied on both datasets, with the null hypothesis corresponding to non-stationarity. Using a 95\% confidence level, 44\% of the series in the proposed dataset were found to be stationary, compared with only 5\% in the M4 dataset. Furthermore, the information entropy is measured at an average of $4.17$ bits for the M4 dataset, while the REAL-V-TSFM dataset shows an average of $3.88$ bits, reflecting a slightly lower degree of variability and uncertainty in the latter. 

\begin{figure}[htbp]
  \centering
  \includegraphics[width=0.40\textwidth]{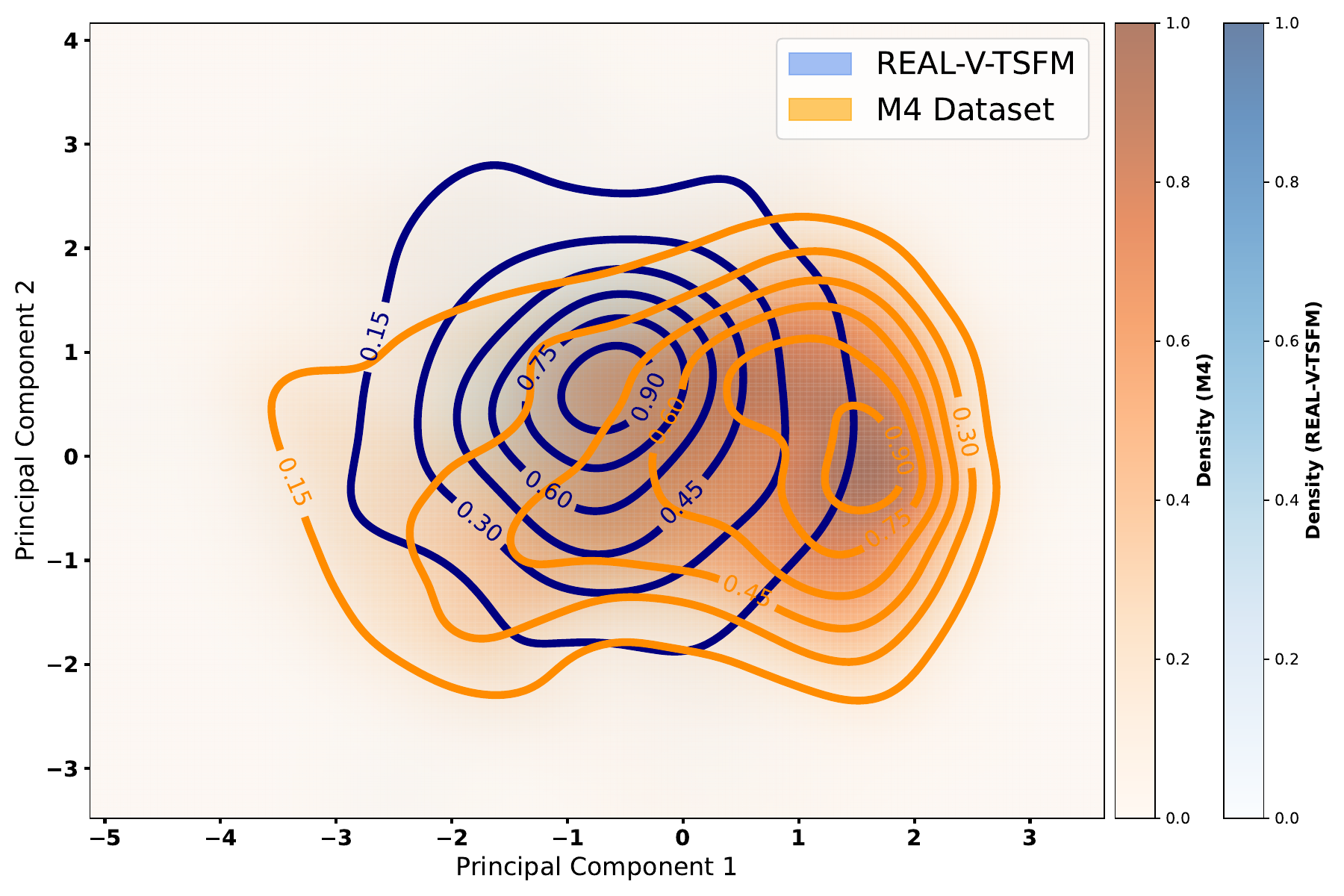}
  \caption{PCA projection of the proposed dataset and the M4-Daily dataset. The color of the heatmap represents the density level, with darker colors indicating regions of higher point concentration. The contour lines (also referred to as iso-density or equal-density curves) connect points sharing the same density value.}
  \label{fig:PCA_prjection}
\end{figure}

Finally, we project both datasets using principal component analysis (PCA) and observe markedly different distributions. We align the two time series datasets REAL-V-TSFM and M4 to the same sequence length and performing min–max normalization on each sequence individually, and then we combined them and applied PCA. We visually depict the distribution shapes, concentration regions, and overlap \& divergence patterns of the two datasets. As illustrated in Fig.~\ref{fig:PCA_prjection}, the two time series datasets demonstrate notable differences in their low-dimensional projections: REAL-V-TSFM is distributed more uniformly, whereas M4 displays a clear skew, with its distribution being sparse in the lower-left and right regions of the PCA plane. By contrast, our time series are derived from more variable real-world physical processes, leading to uniform distributional coverage.

\begin{table*}[!htbp]
\centering
\begin{adjustbox}{max width=0.9\textwidth}
\begin{tabular}{llcccc}
\toprule
\textbf{Model} &
  \multicolumn{1}{c}{\textbf{Datasets}} &
  \textbf{MAPE } &
  \textbf{sMAPE} &
  \textbf{Agg. Relative WQL} &
  \textbf{Agg. Relative MASE} \\
\midrule
\multirow{5}{*}{amazon/chronos-bolt-base} &
  REAL-V-TSFM & \textbf{7.32 ± 17.06} & \textbf{6.57 ± 9.93} & \textbf{0.93 ± 0.90} & \textbf{0.67 ± 0.63} \\
 & M4-Weekly        & 5.72 ± 3.83            & 5.70 ± 3.83            & 0.79 ± 0.87          & 0.50 ± 0.49          \\
 & M4-Daily         & 4.93 ± 3.82            & 5.03 ± 3.22            & \textbf{1.00 ± 0.85} & 0.63 ± 0.59          \\
 & electricity\_D   & 6.52 ± 3.03            & 6.44 ± 2.88            & 0.80 ± 0.41          & 0.62 ± 0.35          \\
 & LOOP\_SEATTLE\_D & \textbf{6.65 ± 2.61}   & \textbf{6.75 ± 2.70}   & 0.90 ± 0.09          & \textbf{0.89 ± 0.09} \\
\midrule
\multirow{5}{*}{amazon/chronos-t5-large} &
  REAL-V-TSFM & \textbf{9.32 ± 18.46} & 8.40 ± 9.69 & \textbf{5.45 ± 31.23} & \textbf{5.58 ± 34.82} \\
 & M4-Weekly        & 8.85 ± 6.17            & \textbf{8.70 ± 6.06}   & 1.19 ± 1.20          & 0.75 ± 0.76          \\
 & M4-Daily         & 7.11 ± 7.02            & 7.18 ± 4.65            & \textbf{1.56 ± 1.26}          & \textbf{0.98 ± 0.88}          \\
 & electricity\_D   & \textbf{9.18 ± 3.85}   & \textbf{8.99 ± 3.44}   & 1.19 ± 0.56          & 0.88 ± 0.50          \\
 & LOOP\_SEATTLE\_D & 7.06 ± 2.83            & 7.18 ± 2.92            & 0.98 ± 0.16          & 0.95 ± 0.13          \\
\midrule
\multirow{5}{*}{google/timesfm-2.0-500m-pytorch} &
  REAL-V-TSFM & \textbf{6.97 ± 16.63} & 6.24 ± 9.17 & \textbf{0.91 ± 1.02} & \textbf{0.64 ± 0.65} \\
 & M4-Weekly        & \textbf{8.30 ± 7.08}            & 8.18 ± 6.70            & 0.85 ± 0.96          & 0.60 ± 0.62          \\
 & M4-Daily         & 1.9 ± 1.08             & 2.03 ± 2.55            & 0.39 ± 0.22          & 0.23 ± 0.25          \\ 
 & electricity\_D   & 6.81 ± 3.63            & \textbf{6.77 ± 3.54}   & 0.80 ± 0.44          & 0.63 ± 0.39          \\
 & LOOP\_SEATTLE\_D & 6.79 ± 2.77            & \textbf{6.89 ± 2.86}   & \textbf{0.92 ± 0.06} & \textbf{0.90 ± 0.07} \\
\midrule
\multirow{5}{*}{LinearRegression} &
  REAL-V-TSFM & \textbf{15.52 ± 28.44} & \textbf{14.28 ± 20.21} & 1.00 ± 0.00 & 1.00 ± 0.00 \\
 & M4-Weekly        & \textbf{21.91 ± 23.49} & \textbf{19.96 ± 19.67} & 1.00 ± 0.00          & 1.00 ± 0.00          \\
 & M4-Daily         & 14.14 ± 24.99          & 14.10 ± 19.99           & 1.00 ± 0.00          & 1.00 ± 0.00          \\
 & electricity\_D   & 13.56 ± 10.74          & 13.93 ± 14.34          & 1.00 ± 0.00          & 1.00 ± 0.00          \\
 & LOOP\_SEATTLE\_D & 7.50 ± 2.97            & 7.63 ± 3.13            & 1.00 ± 0.00          & 1.00 ± 0.00          \\
\bottomrule
\end{tabular}
\end{adjustbox}
\caption{Performance comparison across models and datasets. Boldface values indicate the lowest and second-lowest performance within each group of models. Each item is reported as $\mu \pm \text{std}$, representing the mean and standard deviation, respectively.}
\label{tab:forecast_comparison}
\end{table*}

\section{Testing REAL-V-TSFM}
\label{evaluation_dataset}

\subsection{Evaluation Settings}

Upon building the REAL-V-TSFM dataset, we also evaluated four forecasting datasets from GIFT\_EVAL\cite{aksu2024giftevalbenchmarkgeneraltime}, shown in Table \ref{tab:forecast_comparison}. To ensure consistent input dimensions across time series of varying lengths, all data are segmented into fixed-size windows of 500 time steps, where the first 450 steps serve as the contextual input and the remaining 50 as the prediction horizon. For longer sequences exceeding 500 time steps, we apply a sliding-window approach that scans through the series in strides of 500, effectively partitioning the data into multiple overlapping segments. For shorter sequences (i.e., those with fewer than 500 time steps), we perform linear interpolation to extend them to a uniform length of 500. This procedure guarantees that every time series contributes at least one complete 500-step segment for evaluation, ensuring comprehensive coverage across datasets of diverse lengths.

Model performance is assessed using four complementary metrics: (1) \textbf{Mean Absolute Percentage Error}: MAPE measures the average magnitude of forecasting errors as a percentage of actual values (2) \textbf{Symmetric MAPE}: normalizes errors by the average of actual and predicted values \cite{chicco2021coefficient} (3) \textbf{Aggregate Relative Weighted Quantile Loss}: evaluates the accuracy and calibration of quantile forecasts across multiple series, weighted by their importance, and normalized against a reference baseline \cite{shchur2023autogluon}  and finally, (4) \textbf{Aggregate Relative Mean Absolute Scaled Error}: Aggregate Relative MASE provides a scaled measure of forecast accuracy by comparing model errors to those of a baseline method, aggregated across all series \cite{hyndman2006another}. As baseline, the performance of a Linear Regression model is directly adopted. Three open-source TSFMs from Hugging Face were selected for performance comparison as shown in Table \ref{tab:models_in_main}, with additional models evaluated as detailed in Table \ref{tab:different_size_results}.

\begin{table}[!htbp]
\centering
\begin{adjustbox}{max width=0.47\textwidth}
\begin{tabular}{lcccc}
\toprule
\textbf{Model} & \textbf{Params (M)} & \textbf{Release Date} & \textbf{Architecture} & \textbf{Reference} \\
\midrule
amazon/chronos-bolt-base    & $\sim$205M  & 2024 & Encoder-Decoder    & \cite{ansari2024chronos} \\
\midrule
amazon/chronos-t5-large     & $\sim$709M  & 2024 & Encoder-Decoder & \cite{ansari2024chronos} \\
\midrule
google/timesfm-2.0-500m     & $\sim$500M        & 2025 & Decoder-only    & \cite{das2024decoder} \\
\midrule
LinearRegression            & --          & classical & Linear Model & (Baseline) \\
\bottomrule
\end{tabular}
\end{adjustbox}
\caption{Comparison of evaluated TSFMs}
\label{tab:models_in_main}
\end{table}

\subsection{REAL-V-TSFM is challenging}

As shown in Table \ref{tab:forecast_comparison}, we generally observe that performances on REAL-V-TSFM rank either the first or second worst in terms of forecasting  across nearly all models. This indicates that our dataset is more challenging than other datasets in the GIFT-EVAL benchmark. When using Agg. Relative WQL as the evaluation metric, which emphasizes distributional characteristics, the difference becomes even more prominent: on chronos-t5-large, the performance gap compared with other datasets reaches 5.45, whereas for other models, it is around 1.0. This result demonstrates that current TSFM models fail to adequately capture the predictive distributions of time series data reflecting real physical laws, thereby revealing their limitations in generalizability.

It is worth noting that in terms of model predictability, the performance disparity is not significantly larger than that observed on other GIFT-EVAL datasets. For instance, in the performance results of the google/timesfm-2.0-500m-pytorch model, the Agg. Relative WQL value on the proposed dataset is even better than that on \texttt{LOOP\_SEATTLE\_D}, indirectly confirming the predictability of the model. However, overall, the models do exhibit a certain degree of predictive performance degradation on time series that reflect motion dynamics extracted using optical flow methods.

Regarding model comparison, timesfm-2.0-500m-pytorch demonstrates overall better performance than the chronos series, particularly on our dataset, where it exhibits relatively stronger generalization. This advantage may be attributed to its decoder-only architecture. Moreover, the bolt version shows substantial improvement over the t5 version, with significant gains in both performance and generalization ability. Furthermore, as shown in Table \ref{tab:different_size_results}, models of different sizes within the same series show only limited performance improvements, suggesting that a clear scaling law may not be evident. 

\section{Conclusion}
\label{conclusion}

In this work, we propose a novel time series extraction pipeline from real world videos and introduce an open-source dataset, REAL-V-TSFM, built from public video datasets. Temporal signals are extracted from real-world videos using optical flow to bridge the gap between synthetic benchmarks and real dynamics. Experiments show that while TSFMs perform well on standard datasets, their performance drops on REAL-V-TSFM, revealing a gap between synthetic and real-world data. This underscores the need for data-centric benchmarks that more effectively capture real-world complexity, as well as data augmentation strategies leveraging this pipeline on a tremendously rich collection of real-life videos for TSFMs pretraining.

\section{Discussion and Future Work}

To the best of our knowledge, this paper is the first to introduce a novel and practically valuable benchmark derived from real-world physical dynamics via optical flow, and then apply this methodology to evaluate cutting-edge TSFMs in the zero-shot forecasting setting, demonstrating substantial potential for enriching the diversity of time series datasets, particularly in today’s video-rich online environment. However, despite the promising generalization capabilities of TSFMs, many real-world deployments still rely on task-specific fine-tuning, so using a portion of this proposed dataset as a training set for few-shot prediction would be informative for assessing both the dataset’s utility and TSFMs’ practical generalization. Moreover, incorporating a broader set of other TSFMs as baselines, such as N-BEATS\cite{Oreshkin2019NBEATSNB} and PatchTST \cite{nie2023timeseriesworth64}, would provide a more comprehensive performance landscape. From a theoretical standpoint, developing explanations for why real-world video-derived datasets induce performance degradation in current TSFMs is crucial for understanding model limitations; in addition, although this work adopts a single optical-flow method, alternative or more recent variants could serve as the basis for constructing comparable datasets across different extraction pipelines. Finally, broader evaluations on additional tasks such as imputation and classification, together with more diverse real-world video sources, would further substantiate claims regarding the universality and robustness of TSFMs.

% Future work should incorporate richer modalities, improve model adaptability, and validate results across more diverse datasets.

\section{Acknowledgments}

All models and resources developed in this work are strictly intended for research and educational purposes; no model weights or derivatives are used — or will be used — for any commercial application. We exclusively utilize publicly available corpora or datasets for which explicit authorization has been obtained from the original data providers. All license terms have been reviewed to ensure full compliance with copyright, attribution, and sharing requirements.

All code, models, and processed data artifacts will be released under an open-source, research-oriented license (e.g., CC BY-NC), accompanied by comprehensive documentation and bias-analysis methodology to promote transparency and reproducibility.  We commit to ongoing ethical oversight through periodic reevaluation of datasets and model outputs, prompt updates in response to emerging concerns, and consultation with interdisciplinary advisory boards to ensure adherence to the highest ethical standards.

\appendix

\bibliography{aaai2026}
\section{Models}

As shown in Table~\ref{tab:models}, we mainly select multiple variants of the Chronos models, and we also evaluate the TimesFM models. The key architectural distinction lies in the fact that TimesFM adopts a \textit{decoder-only} design, while the Chronos models are based on an \textit{encoder--decoder} architecture. Their sizes range from the largest at 709M parameters to the smallest at 7M parameters, representing some of the most commonly used TSFMs in recent studies. It is worth noting that, for TimesFM, we only tested the case with frequency $=0$, i.e., the high-frequency setting, since we assume that the video-derived time series used in our experiments are predominantly of high frequency.

\begin{table*}[!htbp]
\centering
\begin{adjustbox}{max width=\textwidth}
\begin{tabular}{lcccc}
\toprule
\textbf{Model} & \textbf{Params (M)} & \textbf{Release Date} & \textbf{Architecture} & \textbf{Reference} \\
\midrule
amazon/chronos-bolt-tiny    & $\sim$7M   & 2024 & Encoder-Decoder    & \cite{ansari2024chronos} \\
amazon/chronos-bolt-mini    & $\sim$21M   & 2024 & Encoder-Decoder    & \cite{ansari2024chronos} \\
amazon/chronos-bolt-small   & $\sim$48M  & 2024 & Encoder-Decoder    & \cite{ansari2024chronos} \\
amazon/chronos-bolt-base    & $\sim$205M  & 2024 & Encoder-Decoder    & \cite{ansari2024chronos} \\
\midrule
amazon/chronos-t5-tiny      & $\sim$8M   & 2024 & Encoder-Decoder & \cite{ansari2024chronos} \\
amazon/chronos-t5-mini      & $\sim$20M   & 2024 & Encoder-Decoder & \cite{ansari2024chronos} \\
amazon/chronos-t5-small     & $\sim$46M   & 2024 & Encoder-Decoder & \cite{ansari2024chronos} \\
amazon/chronos-t5-base      & $\sim$201M  & 2024 & Encoder-Decoder & \cite{ansari2024chronos} \\
amazon/chronos-t5-large     & $\sim$709M  & 2024 & Encoder-Decoder & \cite{ansari2024chronos} \\
\midrule
% google/timesfm-1.0-200m     & $\sim$200M        & 2024 & Decoder-only    & \cite{das2024decoder} \\
google/timesfm-2.0-500m     & $\sim$500M        & 2025 & Decoder-only    & \cite{das2024decoder} \\
\midrule
LinearRegression            & --          & classical & Linear Model & (Baseline) \\
\bottomrule
\end{tabular}
\end{adjustbox}
\caption{Comparison of evaluated TSFMs details}
\label{tab:models}
\end{table*}

\section{Optical Flow Experimental Details}
\label{sec:optical_flow_details}

The \texttt{calcOpticalFlowPyrLK} function from OpenCV is employed, 
with a relatively large search window of $(40 \times 40)$ to improve the tracking of fast-moving objects. A three-level image pyramid is adopted to enhance robustness in scenarios involving large displacements. In addition, the convergence criteria were set to a maximum of 30 iterations or an accuracy threshold of 0.01, thus balancing computational cost and precision. Overall, this configuration provides a robust solution for optical flow estimation in the cases of rapid motion or large displacements, 
albeit at the expense of increased computational complexity.also Forward-backward check is also tested as a quality verification method for optical flow and set the forward-backward error threshold  to 50.0 and the single-direction optical flow residual error threshold (ERR\_THRESH) to 80.0. Increasing FB\_ERR\_THRESH is intended to retain more tracked points, even those with relatively large errors, thereby extending the length of the time series, though inevitably introducing some noise from erroneous tracking. Similarly, the higher ERR\_THRESH allowed for more relaxed error filtering, further increasing the number of valid tracked points and enhancing coverage of the optical flow data at the potential cost of reduced accuracy. In corner detection, OpenCV's \texttt{goodFeaturesToTrack} function is applied, setting the \texttt{maxCorners} parameter to 30 as a trade-off between performance and efficiency, and the \texttt{qualityLevel} parameter to 0.01.

\section{Time Series Example}
\begin{figure*}[htbp]
    \includegraphics[width=1.\linewidth]{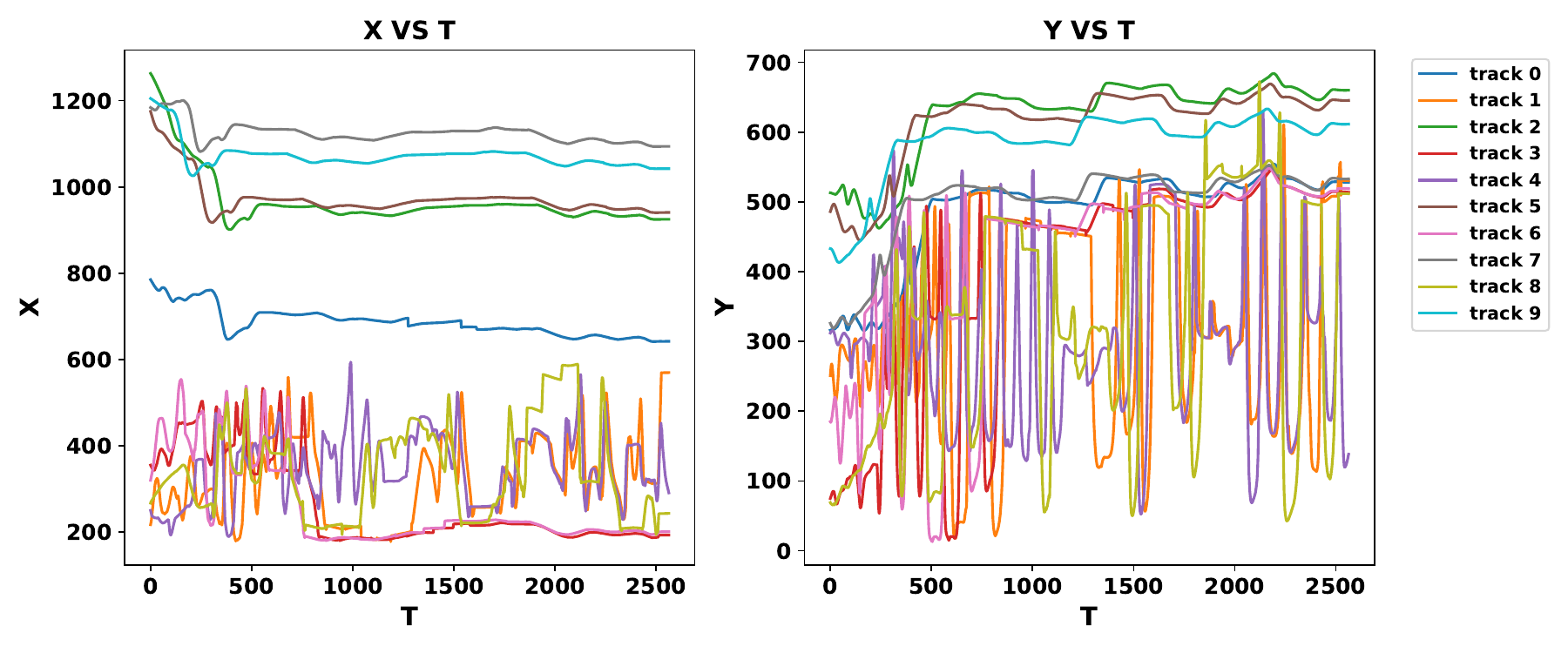}
    \caption{Ten time series data extracted from real video data}
    \label{fig:outpu_max}
\end{figure*}

Extracting time-series information from videos requires meticulous manual verification. Here, we present an illustrative example of the extracted time series, as shown in Figure~\ref{fig:outpu_max}. We display ten object motion trajectories tracked from the video using the optical flow method. It can be observed that the temporal variation patterns along the X and Y axes may differ significantly. Some stationary time series are visible in the graph, such as uniform camera shake caused by human breathing, while there are also nonstationary sequences, which contribute to increasing the diversity of the dataset.

\section{Principal Component Analysis (PCA)}
\label{sec:pca}

The similarity of the REAL-V-TSFM dataset is evaluated in comparison to other datasets, highlighting the enhanced diversity introduced by this newly constructed time series dataset. For both the REAL-V-TSFM and M4 datasets, we first normalize and then merge them prior to performing a PCA. When the two datasets exhibit a closer distribution in the principal projection space, their similarity is considered higher; conversely, the greater the divergence, the lower their similarity. This approach thus reflects whether the datasets reveal comparable structures within the same low-dimensional principal component space.

% \begin{figure}[!htbp]
% \centering
%     \hspace{-1.2cm}
%     \includegraphics[width=0.8\linewidth]{photos/PCA_density_overlays.pdf}
%     \caption{PCA projection for demonstrating the non-similarity between two datasets}
%     \label{fig:PCA_prjection}
% \end{figure}

% \begin{wrapfigure}{R}{0.50\textwidth}
%   \centering
%   \vspace{-2pt}
%   \includegraphics[width=0.50\textwidth]{photos/PCA_density_overlays.pdf}
%   \caption{PCA projection for demonstrating the non-similarity between two datasets}
%   \label{fig:PCA_prjection}
% \end{wrapfigure}

We align the two time series datasets to the same sequence length and performing min–max normalization on each sequence individually, and then we combined them and applied PCA. Two-dimensional kernel density estimates were then computed separately for the projections of each dataset. By overlaying semi-transparent heatmaps with contour lines in blue (M4) and orange (REAL‑V‑TSFM), we visually depict the distribution shapes, concentration regions, and overlap \& divergence patterns of the two datasets in the first two principal component dimensions. As illustrated in Fig.~\ref{fig:PCA_prjection}, the two time series datasets demonstrate notable differences in their low-dimensional projections: REAL-V-TSFM is distributed more uniformly, whereas M4 displays a clear skew, with its distribution being sparse in the lower-left and upper-right regions of the PCA plane. This observation not only underscores the limited diversity of existing time series datasets, but also demonstrates the feasibility and significance of extracting diverse and information-rich time series directly from video data. Given the massive availability of video sources, the potential for extracting varied and valuable time series data is both substantial and promising.

\section{Does model size really matter?}

We evaluated the full range of Chronos models of varying sizes and compared their performance against other models, as shown in Table~\ref{tab:different_size_results}. Our findings suggest that increasing model size leads to only limited performance gains. Moreover, the \textit{decoder-only} TimesFM model demonstrates substantially better performance than other models on the M4 dataset, but its degraded performance on REAL-V-TSFM highlights limitations in its generalization ability. Interestingly, smaller models, such as the \textit{tiny} variant, can achieve performance comparable to or even on par with larger counterparts such as \textit{large} or \textit{base}. Consequently, scaling laws do not appear to consistently hold for TSFMs, and this observation provides valuable guidance for continued exploration in this domain.

\begin{table*}[htbp]
\centering
\begin{adjustbox}{max width=\textwidth}
\begin{tabular}{llcccc}
\toprule
\textbf{Model} & 
\multicolumn{1}{c}{\textbf{Datasets}} & 
\textbf{MAPE} & 
\textbf{sMAPE} & 
\textbf{Agg. Relative WQL} & 
\textbf{Agg. Relative MASE} \\
\midrule
\multirow{5}{*}{amazon/chronos-bolt-tiny} &
REAL-V-TSFM &
\textbf{7.43 ± 17.40} &
\textbf{6.66 ± 9.92} &
\textbf{0.92 ± 0.88} &
\textbf{0.67 ± 0.62} \\
& M4-Weekly &
7.22 ± 5.33 &
7.12 ± 5.27 &
0.91 ± 0.96 &
0.59 ± 0.55 \\
& M4-Daily &
5.07 ± 1.03 &
5.14 ± 4.03 &
\textbf{0.95 ± 0.77} &
0.63 ± 0.59 \\
& electricity\_D &
6.79 ± 3.43 &
\textbf{6.65 ± 3.18} &
0.82 ± 0.41 &
0.64 ± 0.38 \\
& LOOP\_SEATTLE\_D &
\textbf{6.03 ± 2.39} &
6.15 ± 2.49 &
0.82 ± 0.09 &
\textbf{0.81 ± 0.09} \\
\midrule

\multirow{5}{*}{amazon/chronos-bolt-mini} &
REAL-V-TSFM &
\textbf{7.37 ± 17.22} &
6.65 ± 10.16 &
\textbf{0.92 ± 0.90} &
\textbf{0.67 ± 0.63} \\
& M4-Weekly &
6.88 ± 5.01 &
\textbf{6.76 ± 4.90} &
0.90 ± 0.98 &
0.57 ± 0.57 \\
& M4-Daily &
5.01 ± 4.88 &
5.08 ± 4.78 &
\textbf{0.97 ± 0.87} &
0.62 ± 0.62 \\
& electricity\_D &
6.73 ± 3.26 &
6.61 ± 3.05 &
0.83 ± 0.43 &
0.65 ± 0.39 \\
& LOOP\_SEATTLE\_D &
\textbf{6.69 ± 2.63} &
\textbf{6.79 ± 2.73} &
0.91 ± 0.08 &
\textbf{0.90 ± 0.08} \\
\midrule

\multirow{5}{*}{amazon/chronos-bolt-small} &
REAL-V-TSFM &
\textbf{7.50 ± 18.80} &
6.55 ± 9.38 &
\textbf{0.92 ± 0.88} &
\textbf{0.67 ± 0.62} \\
& M4-Weekly &
6.64 ± 4.88 &
6.55 ± 4.86 &
0.87 ± 0.97 &
0.55 ± 0.55 \\
& M4-Daily &
5.00 ± 3.98 &
5.09 ± 3.88 &
\textbf{0.96 ± 0.88} &
0.62 ± 0.45 \\
& electricity\_D &
\textbf{6.71 ± 3.16} &
\textbf{6.60 ± 2.96} &
0.81 ± 0.42 &
0.64 ± 0.38 \\
& LOOP\_SEATTLE\_D &
6.58 ± 2.62 &
\textbf{6.66 ± 2.70} &
0.89 ± 0.09 &
\textbf{0.88 ± 0.09} \\
\midrule

\multirow{5}{*}{amazon/chronos-bolt-base} &
REAL-V-TSFM &
\textbf{7.32 ± 17.06} &
\textbf{6.57 ± 9.93} &
\textbf{0.93 ± 0.90} &
\textbf{0.67 ± 0.63} \\
& M4-Weekly &
5.72 ± 3.83 &
5.70 ± 3.83 &
0.79 ± 0.87 &
0.50 ± 0.49 \\
& M4-Daily &
4.93 ± 3.82 &
5.03 ± 3.22 &
\textbf{1.00 ± 0.85} &
0.63 ± 0.59 \\
& electricity\_D &
6.52 ± 3.03 &
6.44 ± 2.88 &
0.80 ± 0.41 &
0.62 ± 0.35 \\
& LOOP\_SEATTLE\_D &
\textbf{6.65 ± 2.61} &
\textbf{6.75 ± 2.70} &
0.90 ± 0.09 &
\textbf{0.89 ± 0.09} \\
\midrule

\multirow{5}{*}{amazon/chronos-t5-tiny} &
REAL-V-TSFM &
\textbf{10.40 ± 20.13} &
9.28 ± 10.85 &
\textbf{5.40 ± 30.04} &
\textbf{5.37 ± 33.57} \\
& M4-Weekly &
\textbf{10.84 ± 8.94} &
\textbf{10.41 ± 6.81} &
1.32 ± 1.30 &
0.87 ± 0.81 \\
& M4-Daily &
7.43 ± 3.28 &
7.44 ± 4.64 &
0.59 ± 0.89 &
1.00 ± 1.25 \\
& electricity\_D &
10.12 ± 4.05 &
\textbf{9.82 ± 3.82} &
\textbf{1.35 ± 0.64} &
0.97 ± 0.55 \\
& LOOP\_SEATTLE\_D &
7.87 ± 3.17 &
8.09 ± 3.41 &
1.10 ± 0.25 &
\textbf{1.07 ± 0.23} \\
\midrule

\multirow{5}{*}{amazon/chronos-t5-mini} &
REAL-V-TSFM &
9.36 ± 18.15 &
8.66 ± 10.26 &
\textbf{5.55 ± 32.01} &
\textbf{5.46 ± 34.38} \\
& M4-Weekly &
\textbf{9.98 ± 6.57} &
\textbf{9.77 ± 6.25} &
1.22 ± 1.13 &
0.80 ± 0.69 \\
& M4-Daily &
7.13 ± 4.57 &
7.09 ± 6.22 &
\textbf{1.48 ± 1.32} &
0.93 ± 0.75 \\
& electricity\_D &
\textbf{9.69 ± 4.12} &
\textbf{9.53 ± 3.77} &
1.31 ± 0.63 &
0.93 ± 0.54 \\
& LOOP\_SEATTLE\_D &
7.52 ± 3.03 &
7.74 ± 3.27 &
1.03 ± 0.20 &
\textbf{1.01 ± 0.18} \\
\midrule

\multirow{5}{*}{amazon/chronos-t5-small} &
REAL-V-TSFM &
9.76 ± 19.56 &
8.68 ± 10.07 &
\textbf{4.89 ± 30.06} &
\textbf{4.93 ± 32.30} \\
& M4-Weekly &
\textbf{9.96 ± 6.42} &
\textbf{9.70 ± 6.12} &
1.26 ± 1.16 &
0.81 ± 0.70 \\
& M4-Daily &
7.26 ± 6.13 &
7.28 ± 5.22 &
\textbf{1.53 ± 1.25} &
0.97 ± 0.77 \\
& electricity\_D &
\textbf{9.88 ± 3.74} &
\textbf{9.64 ± 3.30} &
1.34 ± 0.67 &
0.97 ± 0.58 \\
& LOOP\_SEATTLE\_D &
7.54 ± 2.88 &
7.66 ± 2.98 &
1.05 ± 0.15 &
\textbf{1.02 ± 0.13} \\
\midrule

\multirow{5}{*}{amazon/chronos-t5-base} &
REAL-V-TSFM &
\textbf{9.56 ± 21.12} &
8.38 ± 9.92 &
\textbf{5.57 ± 32.11} &
\textbf{5.33 ± 33.89} \\
& M4-Weekly &
9.34 ± 6.41 &
\textbf{9.17 ± 6.20} &
1.24 ± 1.23 &
0.79 ± 0.75 \\
& M4-Daily &
7.31 ± 6.28 &
7.34 ± 6.39 &
\textbf{1.55 ± 1.45} &
0.98 ± 0.78 \\
& electricity\_D &
\textbf{9.41 ± 3.47} &
\textbf{9.21 ± 3.24} &
1.25 ± 0.63 &
0.93 ± 0.57 \\
& LOOP\_SEATTLE\_D &
7.45 ± 2.86 &
7.60 ± 2.98 &
1.03 ± 0.17 &
\textbf{1.01 ± 0.15} \\
\midrule

\multirow{5}{*}{amazon/chronos-t5-large} &
REAL-V-TSFM &
\textbf{9.32 ± 18.46} &
8.40 ± 9.69 &
\textbf{5.45 ± 31.23} &
\textbf{5.58 ± 34.82} \\
& M4-Weekly &
8.85 ± 6.17 &
\textbf{8.70 ± 6.06} &
1.19 ± 1.20 &
0.75 ± 0.76 \\
& M4-Daily &
7.11 ± 7.02 &
7.18 ± 4.65 &
\textbf{1.56 ± 1.26} &
\textbf{0.98 ± 0.88} \\
& electricity\_D &
\textbf{9.18 ± 3.85} &
\textbf{8.99 ± 3.44} &
1.19 ± 0.56 &
0.88 ± 0.50 \\
& LOOP\_SEATTLE\_D &
7.06 ± 2.83 &
7.18 ± 2.92 &
0.98 ± 0.16 &
0.95 ± 0.13 \\
\midrule

\multirow{5}{*}{google/timesfm-1.0-200m-pytorch} &
REAL-V-TSFM &
\textbf{58.29 ± 82.96} &
66.11 ± 70.55 &
\textbf{32.20 ± 75.67} &
32.90 ± 79.88 \\
& M4-Weekly &
\textbf{92.12 ± 103.91} &
\textbf{73.04 ± 61.86} &
24.74 ± 61.39 &
\textbf{25.38 ± 68.59} \\
& M4-Daily &
49.19 ± 55.93 &
55.26 ± 65.22 &
20.77 ± 75.33 &
22.55 ± 26.33 \\
& electricity\_D &
36.88 ± 49.90 &
29.84 ± 33.19 &
\textbf{29.84 ± 33.19} &
4.41 ± 17.42 \\
& LOOP\_SEATTLE\_D &
22.93 ± 9.86 &
19.85 ± 7.61 &
3.45 ± 1.22 &
3.09 ± 1.14 \\
\midrule

\multirow{5}{*}{google/timesfm-2.0-500m-pytorch} &
REAL-V-TSFM &
\textbf{6.97 ± 16.63} &
6.24 ± 9.17 &
\textbf{0.91 ± 1.02} &
0.64 ± 0.65 \\
& M4-Weekly &
\textbf{8.30 ± 7.08} &
\textbf{8.18 ± 6.70} &
0.85 ± 0.96 &
0.60 ± 0.62 \\
& M4-Daily &
1.9 ± 1.08 &
2.03 ± 2.55 &
0.39 ± 0.22 &
0.23 ± 0.25 \\
& electricity\_D &
6.81 ± 3.63 &
6.77 ± 3.54 &
0.80 ± 0.44 &
\textbf{0.63 ± 0.39} \\
& LOOP\_SEATTLE\_D &
6.79 ± 2.77 &
\textbf{6.89 ± 2.86} &
\textbf{0.92 ± 0.06} &
\textbf{0.90 ± 0.07} \\
\midrule

\multirow{5}{*}{LinearRegression} &
REAL-V-TSFM &
\textbf{15.52 ± 28.44} &
\textbf{14.10 ± 20.21} &
1.00 ± 0.00 &
1.00 ± 0.00 \\
& M4-Weekly &
\textbf{21.91 ± 23.49} &
\textbf{19.96 ± 19.67} &
1.00 ± 0.00 &
1.00 ± 0.00 \\
& M4-Daily &
14.14 ± 24.99 &
14.6 ± 19.99 &
1.00 ± 0.00 &
1.00 ± 0.00 \\
& electricity\_D &
13.56 ± 10.74 &
13.93 ± 14.34 &
1.00 ± 0.00 &
1.00 ± 0.00 \\
& LOOP\_SEATTLE\_D &
7.50 ± 2.97 &
7.63 ± 3.13 &
1.00 ± 0.00 &
1.00 ± 0.00 \\
\bottomrule
\end{tabular}
\end{adjustbox}
\caption{Performance comparison across datasets and models.}
\label{tab:different_size_results}
\end{table*}

\section{Objects in REAL-V-TSFM}

\begin{figure*}[!htbp]
    \hspace{-0.2cm}
    \includegraphics[width=1.0\linewidth]{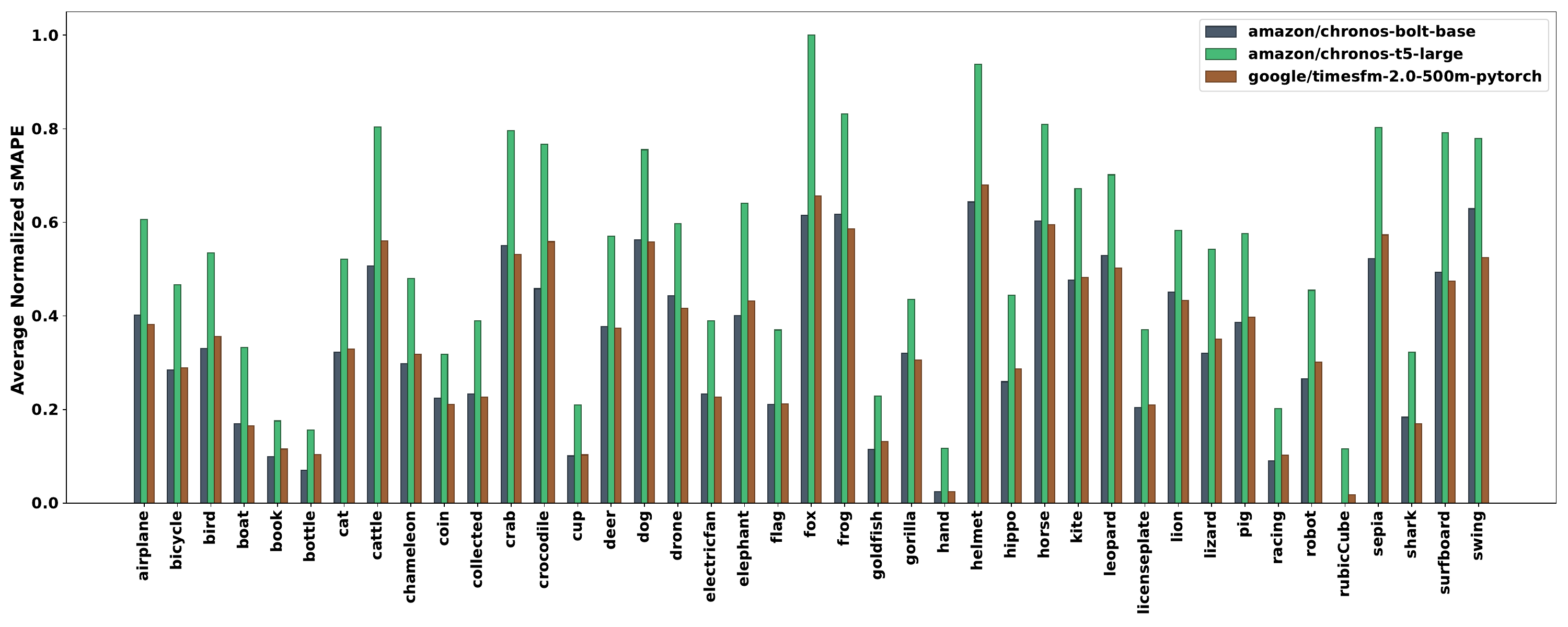}
    \caption{Performance comparison in different objects inside REAL-V-TSFM}
    \label{fig:different_object}
\end{figure*}

Figure~\ref{fig:different_object} illustrates the performance of 3 models on videos containing different objects within the REAL-V-TSFM dataset, where the vertical axis represents the normalized sMAPE. It is evident that the performance varies considerably between different objects, which correspond to diverse types of time series. Interestingly, videos involving animals yield motion-derived time series that are particularly difficult to forecast, resulting in notably lower predictive accuracy. By examining the videos, we observed that animal behaviors often exhibit unpredictable irregularities, particularly in species with a higher degree of freedom in body movement, such as frogs or foxes, which possess four limbs and a movable head. In contrast, species like birds demonstrate more predictable motion patterns during flight, characterized by regular wing flapping, making their motion-derived time series easier for TSFMs to forecast. In contrast, videos of inanimate and static objects, such as books, tend to produce more predictable time series, leading to higher performance. Overall, the dataset exhibits substantial diversity, which not only facilitates the evaluation of model generalization, but also provides a novel and valuable perspective on extracting time series from large-scale real-world data.

\end{document}